\documentclass[a4, conference]{IEEEtran}
\IEEEoverridecommandlockouts
\usepackage{cite}
\usepackage{amsmath,amssymb,amsfonts}
\usepackage{algorithmic}
\usepackage{graphicx}
\usepackage{textcomp}
\usepackage{xcolor}
\usepackage{url}
\usepackage{hyperref}
\def\BibTeX{{\rm B\kern-.05em{\sc i\kern-.025em b}\kern-.08em
    T\kern-.1667em\lower.7ex\hbox{E}\kern-.125emX}}
\begin{document}

\title{Social Robot Navigation with Adaptive Proxemics Based on Emotions

\author{Baris Bilen$^{1}$, Hasan Kivrak$^{2}$, Pinar Uluer$^{3}$, Hatice Kose$^{1}$

\thanks{}}
\thanks{$^{1}$Dept. of AI and Data Engineering, Istanbul Technical University, TURKEY
        {\tt\small \{bilenb20, hatice.kose\}@itu.edu.tr}}%
        
\thanks{$^{2}$Dept. of Computer Engineering, Karabuk University, TURKEY
        {\tt\small kivrakh@karabuk.edu.tr}}%
        
\thanks{$^{3}$Dept. of Computer Engineering, Galatasaray University, TURKEY
        {\tt\small puluer@gsu.edu.tr}}%
}

\maketitle

\begin{abstract}The primary aim of this paper is to investigate the integration of emotions into the social navigation framework to analyse its effect on both navigation and human physiological safety and comfort. The proposed framework uses leg detection to find the whereabouts of people and computes adaptive proxemic zones based on their emotional state. We designed several case studies in a simulated environment and examined 3 different emotions; positive (happy), neutral and negative (angry). A survey study was conducted with 70 participants to explore their impressions about the navigation of the robot and compare the human safety and comfort measurements results. Both survey and simulation results showed that integrating emotions into proxemic zones has a significant effect on the physical safety of a human. The results revealed that when a person is angry, the robot is expected to navigate further than the standard distance to support his/her physiological comfort and safety. The results also showed that reducing the navigation distance is not preferred when a person is happy. 
\end{abstract}


\section{Introduction} 

Socially aware navigation is an active and expanding topic of study that brings together human-robot interaction, perception, and motion planning topics. It is crucial for mobile robots to be able to navigate in a social environment without endangering humans' physical safety. In order to perform and navigate in human-inhabited areas, a mobile robot needs to consider humans not only as an obstacle but as social beings. For this reason, a socially-aware navigation approach that integrates social norms and cues is needed. Most of the state-of-the-art robot navigation algorithms perform inherently in human-populated environments while assuming that all humans are in the same state of mind. However, emotions have a significant impact on our behaviours in our daily lives. Different emotions changes our boundaries in the physical movement space \cite{b1}. Therefore, emotion-aware navigation strategies might play an important role for mobile robots to improve their behaviours while considering human safety and comfort.

In this paper, we designed and developed a social navigation framework for mobile robots to navigate in a human-inhabited environment while interpreting human emotions and integrating them into the decision mechanism of human proxemics. We evaluated the effect of this system and also a survey is conducted with 70 participants. The results show the significant effect of integrating emotions into the decision mechanism of robot during its navigation. 

\section{Related Work}

Robot navigation with social awareness is beneficent for assisting mobile robots in generating socially acceptable behaviours in social environments. To create socially acceptable behaviours, one has to understand the behaviours of humans in a social environment. Edward T. Hall, et al. proposed proxemic criteria for humans defining 4 zones: (i) intimate, (ii) personal, (iii) social and (iv) public \cite{hall_proxemics}. For a robot to have socially acceptable behaviours, it has to avoid entering the personal space of a human. Robot navigation methods that allow a mobile robot to approach people have been thoroughly researched in past years. There are algorithms that widely used for path planning on a static environment such as Dijkstra, A*, D* \cite{d_star_algo}, D* Lite \cite{d_star_light_algo} and RRT* \cite{rrt_star_algo}. Also, the aforementioned methods are studied for the static and dynamic environment but none of them manages to achieve social navigation. A mobile robot may approach a person who is standing, moving, or sitting using conventional human-approaching frameworks. The dynamic window approach (DWA) \cite{dwa} offers a collision avoidance mechanism for mobile robots in dynamic environments. The DWA selects the best way to approach or pass humans by deriving an approximation of the robots' and humans' trajectories. The researchers investigated the preference of people on how to be approached while sitting and presented a cost-based path planning strategy to mimic these preferences \cite{seated_person}. In another study, the authors proposed a framework that allows a mobile robot to use spatial information to establish and sustain a dialogue with standing individuals \cite{spatial_formation}. These frameworks are mostly designed to approach a single person. Although they can be used to avoid obstacles and create collision-free trajectories in human-inhabited environments, they don't offer a physical safety for humans. 

One other way to approach humans is using proxemics theory-based costmaps. Robot operating system (ROS) \cite{ros} offers layered costmap plugins and allow users to work with separate layers such as static map, obstacle, inflation, and other user-specified layers (e.g., socially costmap, range sensor). This plugin can subscribe to objects and people and then alters the costmap by adding Gaussian costs around the subscribed objects and people. By the use of these additional costs, robots can navigate around people in collision-free trajectories by respecting their personal zones \cite{layered_costmap}. The social force model (SFM) \cite{sfm} is another method to describe the motions of pedestrians. SFM is not just used for foreseeing the behaviours of pedestrians but also can be used as a local planner \cite{sfm_as_local_planner}. 

One other approach proposes a learning scheme that can learn the navigational behaviours of people by observing them. This approach has the benefit of adapting to different conditions and increasing its efficiency over time simply by learning on observations. This approach attempts to use human behaviour model on the local planner. Deep reinforcement learning is used in \cite{dqn_rl} for a mobile robot to learn human navigational behaviour using DQN. \cite{a3c_rl} uses  actor-critic method for reinforcement learning to create a model for mobile robot. \cite{elastic_band} extents the time elastic band approach by predicting human behaviours and using this information to create a social navigation planner. Furthermore, people's mood and emotional status can affect their behaviours. For example, in a negative mood, baseline walking speed of pedestrians is likely to decrease \cite{Emotional_wellbeing}. This can also affect their preferred distance in a proxemic zone depending on which mood they are on, i.e., personal space of a person might decrease or increase depending on their mood \cite{b1}. There are several studies involving the use of human emotions into the robot's proxemics in social navigation literature. One of them \cite{emotion_navigation} uses emotion estimation from both faces and trajectories of pedestrians to create a trajectory for a humanoid robot. \cite{dynamic_proxemics} proposes an adjustable proxemic zone depending on people's mood. Although this is a similar work to our work, we also conducted a more comprehensive survey to investigate the adaptive proxemic zones depending on emotions between human-human and human-robot interaction, as well as the expectations for children and the elderly in the same situations.

\begin{figure}[htbp]
\centerline{\includegraphics[width=1 \columnwidth]{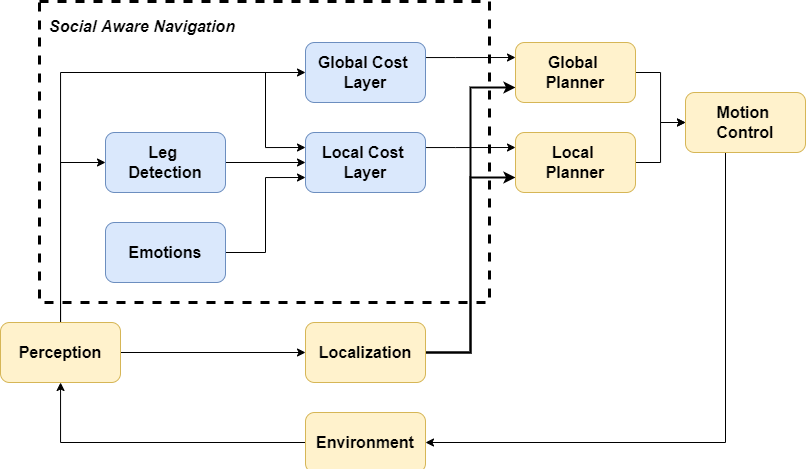}}
\caption{Navigation framework scheme for mobile robots. Consists of 2 parts; 1) Social aware navigation framework and 2) conventional navigation framework.}
\label{framework}
\end{figure}

\section{Framework}

This paper proposes a framework that can integrate adaptive social proxemics and emotions into the navigation framework to see the effects of emotions on a mobile robot's movements and also to analyse its effects on physical safety of humans. For this to be achieved, the robot needs to locate humans in the environment. Figure \ref{framework} shows the architecture of the proposed navigation system. The system consists of two parts: 1) conventional navigation framework and 2) socially aware navigation framework. In the first part, the framework consists of perception, localization, planners (global and local) and motion control function blocks. The second part consists of leg detection, emotion and costmap layer (global and local) function blocks.

\subsection{Leg Detection}

In this paper, a leg detection system \cite{leg_detector} is used. The basic idea of this system is to use laser scan info to determine if there is a human leg nearby. The algorithm processes the laser scan data with machine learning methods to inform about possible legs.

\subsection{Emotions}

We use emotions to increase or reduce the diameter of social proxemic zone depending on the people's emotions. 3 different emotions has been used: 1) Happy, 2) Angry and 3) Neutral. First two emotions are selected from the far edges of valence axis having high arousal components, to be able to observe the clear effects that emotions have on the proxemics and human safety. Selected personal proxemic zone distances for emotions are as follows: for a person with happy emotions, it is 0.5 m, for a person with neutral emotions, it is 1 m \cite{hall_proxemics} and for a person with angry emotions, it is 1.5 m.
 
\subsection{Costmap Layers}

In this study, 4 different layers have been used to generate a costmap: 1) Static layer, 2) Obstacle layer, 3) Inflation layer, and 4) Social layer. The static layer represents the unchanged parts of the costmap. The obstacle layer is used to determine and mark the obstacles as read by laser scan. The inflation layer optimizes or adds costs to important or valuable objects to represent a more realistic costmap for the robot to use. And finally, the social layer tracks the whereabouts of the pedestrians as a layer. This layer also represents the personal space  specifications of pedestrians i.e., the boundaries of the proxemic zone that the robot should not cross over.

\subsection{Global and Local Costmap}

Global costmap uses static, obstacle, and inflation layers to create a global costmap for the robot to navigate in. This costmap is used by the global planner. Local costmap uses all 4 layers to create a local costmap. This costmap is generated in the immediate vicinity of the robot and used by the local planner.

\subsection{Global and Local Planner}

Global planner is used to generating a global path for the robot to track. The global planner uses Dijkstra's algorithm to find the shortest path to the goal while considering the global costmap values. Although the robot can follow the global planner and reach its goal, the local planner ensures that the movements of the robot are within the range of its motors and wheels to prevent any collision or breakdown. Also, it reacts to the new obstacles that might not be there when the global planner generates the path. A good example would be while the robot is following the global path, a new person is detected on the path or near the path and the robot can react to this situation with the help of local planner.

\subsection{Human Safety and Comfort Indices}

To validate our work, we used the social individual index (SII) \cite{hsci}. SII is based on Hall's \cite{hall_proxemics} proxemics. SII is being used to measure the physical safety of each individual. According to SII, physical safety of humans gets in danger if the distance between a robot and a human is smaller than the sum of the area they occupy in the space. Physiological safety is violated if the distance between human and robot is lower than the personal space distance.

\section{Experimental Studies and Results}

In this paper, the proposed framework is simulated using ROS \cite{ros} and Gazebo \cite{gazebo} simulation environments.

\begin{figure}[!h]
\centerline{\includegraphics[width=0.8\columnwidth]{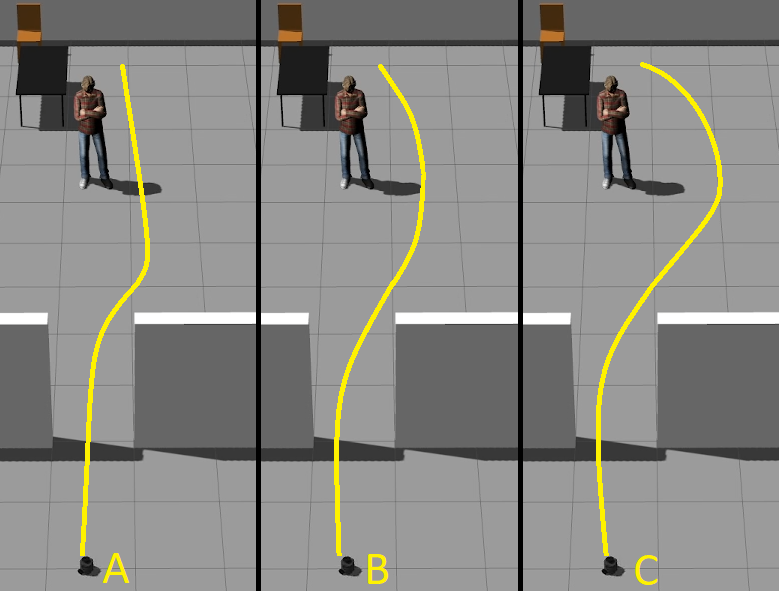}}
\caption{From left to right movement paths of the robot when the person is Happy (A), Neutral (B) or Angry (C)}
\label{paths}
\end{figure}

\subsection{Simulation Study and Results}

In this study, we aim to examine the effect created by knowing a person's emotions and adapting proxemics accordingly. A total of 4 different simulations were performed for 2 different emotions. 2 of the simulations were conducted with a person that has known and unknown happy emotions (see Figure \ref{paths} Robot A and B). Other 2 simulations were conducted with a person that has known and unknown angry emotions (see Figure \ref{paths} Robot C and B).
SII  metric that is used to evaluate these  different simulated actions the robot takes according to the emotions of people, in terms of the people's safety and comfort. We also show the visuals of these simulated actions to the participants and ask them to evaluate the people's safety and comfort subjectively, as reported in the Section \ref{ssc:posttest}.

\begin{figure}[htbp]
\centerline{\includegraphics[width=1\columnwidth]{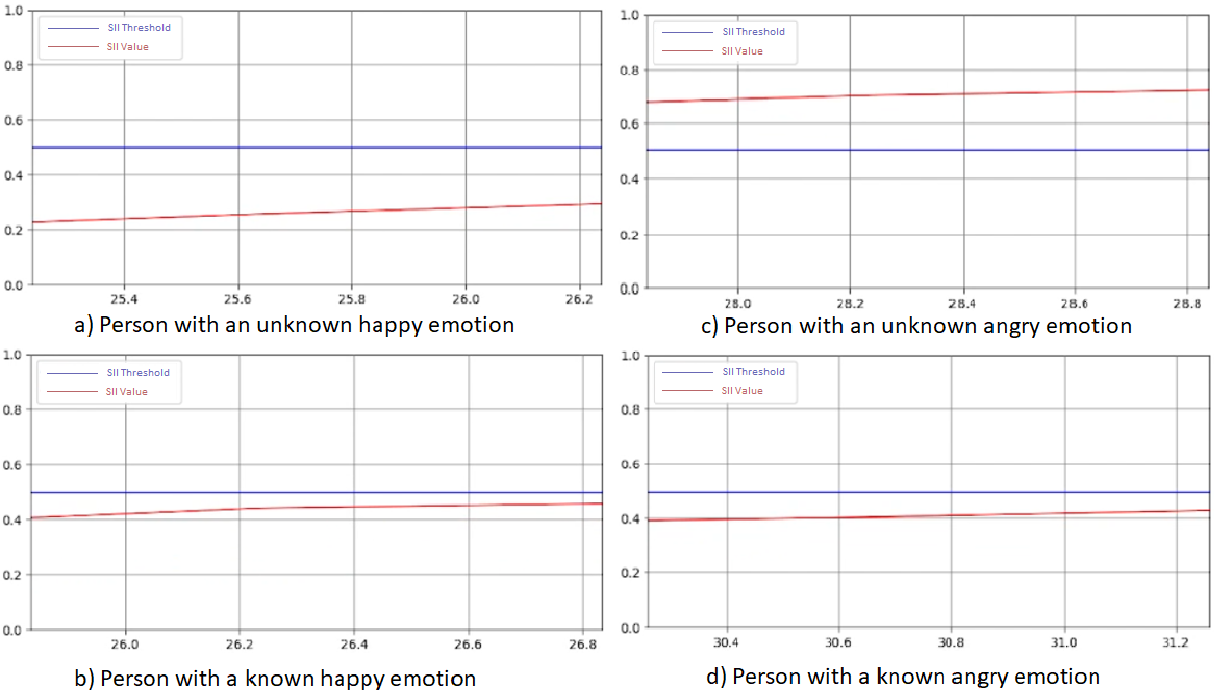}}
\caption{These plots show the physiological safety of a person within different known and unknown emotions. Blue lines represent SII threshold value and red lines represents measured SII value. }
\label{both_emotions}
\end{figure}

\subsubsection{Happy Emotion Simulations}

In happy emotion simulations, we investigated the path created by the robot with known and unknown happy emotion. In the first simulation (See Figure \ref{both_emotions}.a) we decrease the personal proxemic zones diameter of the pedestrian to 0.5 m to simulate happy emotion but did not allow the robot to adjust its proxemic zone accordingly. We observed that the robot passed through the person without exceeding physiological safety. In the second simulation (See Figure \ref{both_emotions}.b) we let the robot adjust its proxemic zone according to the emotion of the person and observed that the robot still did not exceed physiological safety. In the first simulation the total path travelled by the robot was 6 m. and in the second simulation the total path travelled by the robot was 5.5 m. We observed that the detection of the emotion of happiness has a positive effect on the path travelled by the robot.

\subsubsection{Angry Emotion Simulations}

In angry emotion simulations, we investigated the approach distance of a robot passing by a known and an unknown angry person and its effects on the person's physiological safety. To accomplish this, in the first simulation (See Figure \ref{both_emotions}.c) we increase the personal proxemic zone diameter of the pedestrian to simulate angry emotion but did not allow the robot to adjust its proxemics and in the second simulation (See Figure \ref{both_emotions}.d) we let the robot adjust its proxemic zone according to a known angry person. After the simulations, we observed that the detection of angry emotion makes the robot navigate better in terms of physiological safety of humans.

\subsection{NARS and Post-Test Questionnaire Results}

The survey study was conducted with 70 people (34 female, 36 male). The age distribution of the test participants as follows, 12-18 is 4.2\%, 18-30 is 47.9\%, 30-45 is 14.1\%, 45-65 is 25.4\%, 65+ is 8.5\%. Also, 95,8\% of the people has a bachelor or a higher education degree.

\begin{table}[htbp]
\centering
\caption{The Results of One Sample T-Test Based On NARS}
\label{nars}
\begin{tabular}{lllll}
\hline
No. & Subscale & Mean (SD)   & t(69)     & p values        \\ \hline
Q1  & S2       & 3.36 (0.99) & 3.01  & 0.003           \\
Q3  & S3       & 3.10 (1.02) & 0.81  & 0.416           \\
Q4  & S1       & 2.07 (0.95) & -8.02 & \textless 0.001 \\
Q5  & S3       & 3.26 (1.04) & 2.05  & 0.043           \\
Q6  & S3       & 3.17 (0.97) & 1.46  & 0.146           \\
Q8  & S1       & 2.51 (1.04) & -3.89 & \textless 0.001 \\
Q10 & S1       & 2.98 (1.16) & -0.10 & 0.918           \\
Q11 & S2       & 2.53 (1.02) & -3.71 & \textless 0.001 \\
Q12 & S1       & 2.21 (0.97) & -6.61 & \textless 0.001 \\
Q13 & S2       & 2.26 (1.00) & -6.08 & \textless 0.001 \\
Q14 & S2       & 2.68 (1.06) & -2.45 & 0.016           \\ \hline
\end{tabular}
\end{table}

\subsubsection{Negative Attitudes Toward Robots Scale (NARS)} 

We used NARS to evaluate participants' attitudes toward robots, i.e. to find out if they have any prejudice towards robots \cite{nars}. NARS is commonly used to evaluate the attitudes of participants in human-robot interaction and to explain the behavioural disparities between them. NARS has 14 items, categorized in 3 sub-scales: S1 (6 items), S2 (5 items), S3 (3 items). S1 is to measure the negative attitude towards interaction with robots, S2 is to measure the negative attitude toward the social influence of robots and S3 is to measure the negative attitude toward emotional interactions with robots. Each item is scored based on a 5-point semantic differential scale (1 being strongly disagree and 5 being strongly agree).

A one-sample t-test is performed for all items shown in Table \ref{nars} based on the participants' answers. Item numbers are written to be equivalent to the item numbers specified in the NARS test.  The results showed that participants did not have any priors about the robots, we conducted a one-tailed t-test for each item to see the difference in the corresponding means. Table \ref{nars} shows the items with their statistics.

The t-test results showed that all items from S1 sub-scale were smaller means and all items from S3 were higher means which was the expected outcome and showed us that participants has positive tendency towards robots. The results showed most of the S2 sub-scale questions means were smaller which was also expected. Even though the average of the Q1 questions had higher means, in another question (``Do you think our emotions affect interactions with robots?") 62\% of the participants stated that their emotions are effective in robot interaction.

\subsubsection{Post-Test Form}
\label{ssc:posttest}

It is a basic form that includes demographic questions and additional 7 inquiries (Table \ref{post-test_questions}) scored on a same semantic differential scale with NARS. Simulated robot behaviours in Figure \ref{paths} are also used in the questionnaire. First 2 inquires (I1 and I2) are asked to see that whether the theory presented was also valid among humans. The one sample t-test showed (see Table \ref{post-test_results}) that the theory presented is valid for I2. Then, simulation videos and robot behaviours in Figure \ref{paths} are shown to the participants and they were asked to answer the remaining inquires according to this video and picture. We also conducted a one sample t-test to the rest of the inquires, which is also shown in Table \ref{post-test_results}. The results of I3 showed that people find the movement of the robot acceptable when a person is in neutral emotion. The I4 and I6 inquiries results showed that participants prefer robots to pass further away when they are angry but interestingly, there is uncertainty among the participants for robots passing closer when a person is happy. Another point, even though the NARS results showed that the participants did not have any negative attitudes towards robot, the results for the inquiries I5 and I7 showed that if the person to be passed is a child or an old person, they  preferred the robot to pass from a further distance.

\begin{table}[]
\centering
\caption{Post-Test Form Questions}
\label{post-test_questions}
\resizebox{\columnwidth}{!}{%
\begin{tabular}{ll}
\hline
No. & Inquiries                                                                                   \\ \hline
I1  & It doesn't bother me when people I don't know stand closer to me than usual when I'm happy. \\
I2  & I prefer people to stay at a further distance from me than usual when I am angry.          \\
I3  & The distance between the human and the robot B was sufficient as the robot passed the human. \\
I4  & In the picture, the human is disturbed by the distance the robot A has passed.             \\
I5  & Robot A's behavior would be correct if the person in the picture was elder or a child.        \\
I6  & In the picture, the human is disturbed by the distance traveled by the robot C.            \\
I7 & Robot C's behavior would be correct if the person in the picture was elder or a child       
\end{tabular}%
}
\end{table}

\begin{table}[]
\centering
\caption{The Results of One Sample T-Test Based On Post-Test Form}
\label{post-test_results}
\begin{tabular}{llll}
\hline
No. & Mean (SD)   & t(69)     & p values        \\ \hline
I1  & 3.14 (1.05) & 1.13  & 0.260           \\
I2  & 3.98 (0.75) & 10.81 & \textless 0.001 \\
I3  & 3.76 (0.80) & 7.90  & \textless 0.001 \\
I4  & 3.02 (1.10) & 0.21  & 0.829           \\
I5  & 2.44 (1.00) & -4.53 & \textless 0.001 \\
I6  & 2.18 (0.68) & -9.75 & \textless 0.001 \\
I7 & 3.42 (0.84) & 4.12  & \textless 0.001
\end{tabular}
\end{table}

\section{Conclusion}

This paper presents a socially aware navigation framework that can adjust its proxemic distance depending on a person's emotions to maintain people's physical safety.In this work, emotions of the pedestrians are used to change their personal space during the social navigation task of the robot. This paper investigates the effects of integrating emotions in robot's social navigation on the comfort and safety of humans. The SII metric and questionnaires  based on several simulated robot actions are used to evaluate these effects. The results showed that detecting emotions and adjusting personal space depending on the human emotion can create a more comfortable and safer environment for humans.

Although the proposed framework can navigate in a human-inhabited environment while respecting human physical safety, a few changes can be made to make the navigation process more efficient, such as employing cameras to detect the pedestrians and their emotions as well as the laser scan data currently being in use.

This work is aimed at safe and comfortable navigation of Pepper robot among children with hearing impairments in hospitals and audiology centres. In the future, we will further improve the proposed framework by adding a real-time emotion detection module, and employ the whole framework on a physical  Pepper humanoid robot to be used in real world human inhabited environments.

\end{document}